%% file: main.tex
\documentclass{article}
\usepackage[preprint,nonatbib]{neurips_2026}
\input{preamble}

\title{Brain-Aligned Latent Actions:\\Imposing Premotor Cortex Geometry on a Robot's Action Vocabulary}
\author{%
  Amit Jadhav\\
  Purdue University Fort Wayne\\
  \texttt{jadhah01@purdue.edu}
}

\begin{document}
\maketitle
\vspace{-2.2em}
\begin{center}
  \includegraphics[width=\textwidth]{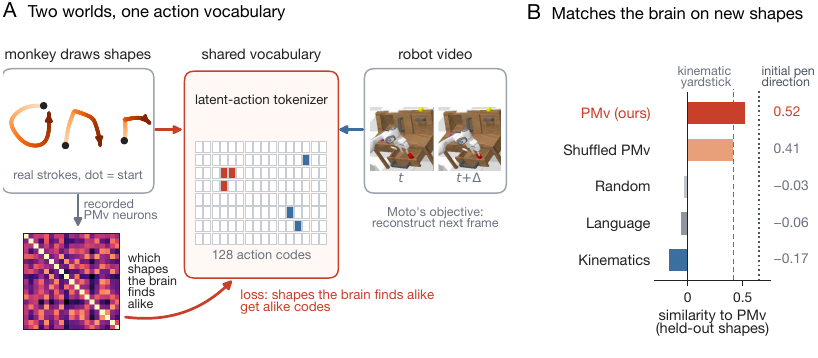}
  \vspace{-1.4em}
  \input{captions/fig1}
  \label{fig:teaser}
\end{center}
\vspace{-0.7em}

\input{sections/00_abstract}

\input{sections/01_intro}
\input{sections/03_vocabulary}
\input{sections/04_method}
\input{sections/05_results}
\input{sections/06_robot}
\input{sections/02_related}
\input{sections/07_limitations}
\input{sections/08_conclusion}

\bibliographystyle{unsrtnat}
\bibliography{refs}

\clearpage
\appendix
\input{sections/A_appendix}

\end{document}

%% file: preamble.tex
\usepackage[T1]{fontenc}
\usepackage[utf8]{inputenc}
\usepackage[scaled=0.9]{inconsolata}
\usepackage{microtype}
\usepackage{amsmath,amssymb}
\usepackage{graphicx}
\usepackage{booktabs}
\usepackage{tabularx}
\newcolumntype{Y}{>{\raggedright\arraybackslash}X}
\usepackage{multirow}
\usepackage{makecell}
\usepackage{xcolor}
\usepackage{xspace}
\usepackage{enumitem}
\setlist{nosep,leftmargin=1.4em}
\usepackage[numbers,sort&compress]{natbib}
\usepackage[font=small,labelfont=bf,skip=4pt]{caption}
\usepackage{subcaption}
\usepackage[hidelinks]{hyperref}
\usepackage[capitalise,noabbrev]{cleveref}

\definecolor{pmvc}{HTML}{C8402A}
\definecolor{shufc}{HTML}{E8A07A}
\definecolor{kinc}{HTML}{3B6FA0}
\definecolor{mutedc}{HTML}{6B7280}

\newcommand{\para}[1]{\vspace{0.15em}\noindent\textbf{#1}\hspace{0.45em}}

%% file: captions/fig1.tex

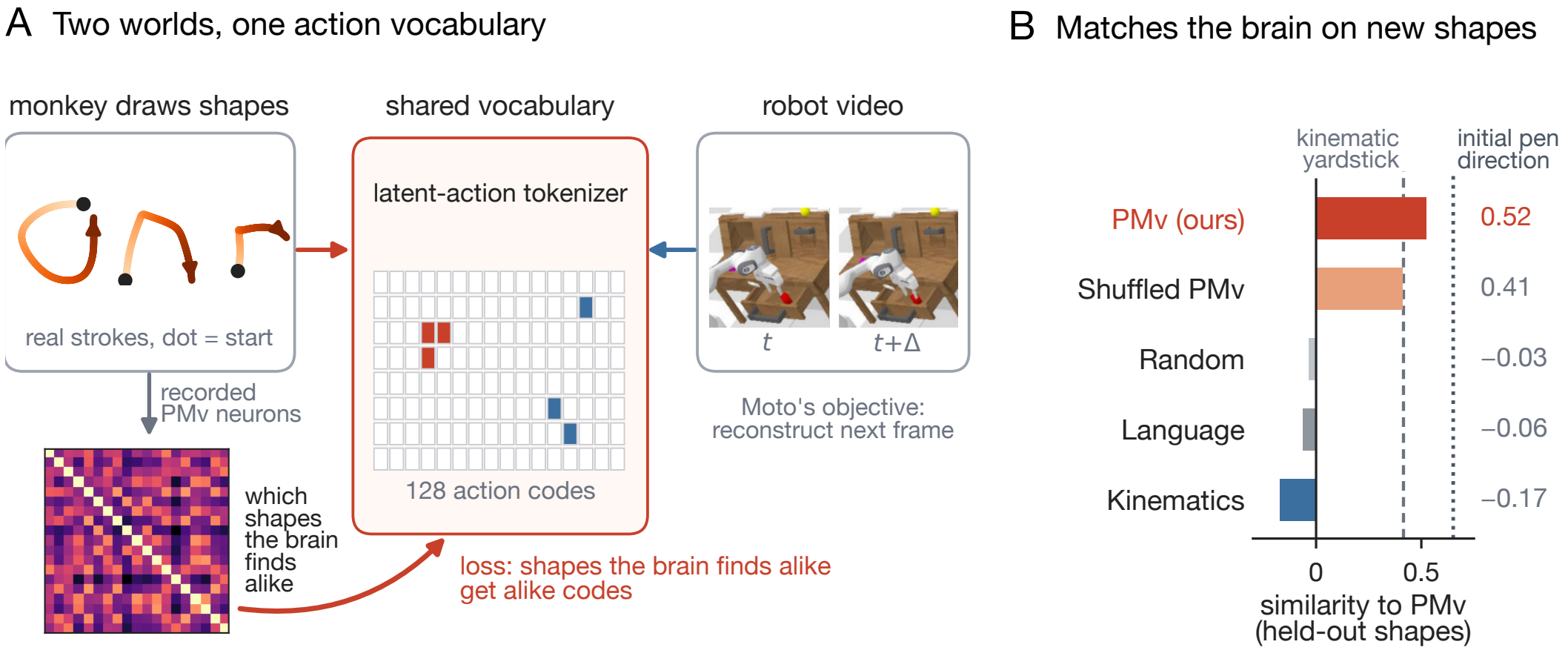
\captionof{figure}{\textbf{Brain-aligned latent actions.} \textbf{(A)}~Moto's latent-action tokenizer encodes robot video into 128 discrete codes, trained as usual to reconstruct the next frame. It also encodes shapes drawn by a macaque (one typical recorded stroke per shape; dot: start), and a loss pulls the geometry of its code use toward the geometry recorded in ventral premotor cortex (PMv). Highlighted codes are illustrative. \textbf{(B)}~On five held-out shape categories, only the PMv-trained tokenizer exceeds a kinematic yardstick (dashed), but an untrained descriptor, the stroke's initial pen direction, scores higher (dotted; $0.65$).}

%% file: sections/00_abstract.tex
\begin{abstract}
\noindent Robot policies increasingly act through a small vocabulary of discrete latent actions learned from unlabeled video, yet training leaves open which motion receives which code. We find this choice highly unstable: two independently trained runs of Moto's latent motion tokenizer assign the same robot clips to nearly unrelated codes (adjusted Rand index $0.001$), and evaluating one tokenizer in bf16 instead of fp32 reassigns 74\% of its tokens. We use this freedom to impose the representational geometry of macaque ventral premotor cortex (PMv), recorded while a monkey drew shapes, on the tokenizer's code assignment, with a straight-through similarity loss trained alongside its reconstruction objective on robot video. In one training run per target, on five shape categories the loss never saw, the PMv-trained tokenizer reaches a representational similarity of $0.52$ with PMv, against $0.20$--$0.26$ for released and no-prior tokenizers and $0.42$ for the 95th percentile of random kinematic encoders. Kinematic, language, and random targets do not transfer ($-0.17$ to $-0.03$). Two results limit the claim: a hand-picked kinematic descriptor, the stroke's initial direction, matches PMv better ($0.65$), and a label-shuffled PMv control scores $0.41$. We specify the downstream manipulation test, on a CALVIN pipeline that reproduces Moto's published results, before running it.
\end{abstract}

%% file: sections/01_intro.tex
\section{Introduction}
\label{sec:intro}

Robot foundation models increasingly learn to act from video that carries no action labels. A widely used recipe compresses the change between two frames into a few discrete \emph{latent actions}: tokens from a learned vocabulary that a policy predicts on the way to motor commands~\cite{bruce2024genie,ye2025latent,chen2025moto,bu2025learning}. Because the vocabulary sits on the policy's inference path, the way it groups motions decides what the policy treats as the same action and what it treats as different.

Nothing in the training objective fixes that grouping, and we find that it varies greatly between runs. Two independently trained runs of the Moto latent motion tokenizer~\cite{chen2025moto} assign the same 10{,}240 tokens, from 1{,}280 robot clips, to nearly unrelated codes (adjusted Rand index $0.001$, seed A after 150k steps and seed B after 99k; \cref{sec:vocab}). Evaluating a single trained tokenizer in bf16 instead of fp32 reassigns 74\% of its tokens (\cref{sec:vocab}). The objective decides how well frames are reconstructed, not which motion receives which token.

A design choice that the data leave open is an opening for a prior. In vision, imposing measured brain structure on artificial networks has repeatedly paid off. A fixed model of primary visual cortex at the front of a convolutional network improves its robustness~\cite{dapello2020simulating}; regularizing a network toward the representational similarity of mouse V1 makes it more robust~\cite{li2019learning}; aligning a network to macaque inferior temporal cortex improves both its agreement with human behavior and its adversarial robustness~\cite{dapello2023aligning}. These works did not wait for brain-like structure to emerge; they imposed it. On the action side, this approach has only begun to be explored~\cite{kapitonova2026distilling}, and it has not been applied to the discrete action vocabularies that robot models learn from video. The primate motor system offers a natural target: ventral premotor cortex (PMv) represents drawn shapes as reusable, categorical action symbols~\cite{tian2026neural}, the role that a latent-action vocabulary plays for a robot.

We therefore ask whether \emph{premotor cortex can choose a robot's action vocabulary}. We impose the representational geometry of macaque PMv, recorded while a monkey drew shapes~\cite{tian2026neural}, on the code assignment of the Moto tokenizer (\cref{fig:teaser}A). The tokenizer keeps its original reconstruction objective on robot video. On alternating batches it also encodes the monkey's strokes, and a straight-through similarity loss pulls the geometry of its code usage toward the PMv geometry of each recording session. The loss acts on \emph{which code each input selects}, not on continuous features, because the discrete choice is all the policy sees.

We built the evaluation around controls. Alignment is measured only on stroke categories that the loss never saw. Four control tokenizers are trained identically toward kinematic, language, label-shuffled PMv, and spectrum-matched random targets. We measure alignment with representational similarity analysis (RSA): the rank correlation between the tokenizer's and PMv's dissimilarity structure over the same shape categories. Every score is compared with a kinematic yardstick: the 95th percentile of 100 random nonlinear encoders of the pen's motion, scored the same way on the same held-out categories (\cref{sec:eval}).

\paragraph{Contributions.}
\begin{itemize}
  \item \textbf{The vocabulary is unstable.} Seed and numerical precision each reassign most tokens of a latent-action tokenizer while reconstruction stays good, so the objective does not pin down the assignment of motions to codes (\cref{sec:vocab}).
  \item \textbf{A method to impose neural geometry on a discrete action vocabulary.} A dual-stream objective aligns the geometry of the code \emph{assignment} to per-session neural similarity, across two domains that share no inputs: monkey drawings and robot video (\cref{sec:method}).
  \item \textbf{Premotor geometry reaches held-out shapes; non-neural targets do not.} In one training run per target, the PMv tokenizer reaches $0.52$ held-out RSA with PMv, above the released tokenizer ($0.20$), the two no-prior runs ($0.24$ and $0.26$), and all 100 random kinematic encoders (maximum $0.43$), while kinematic, language, and random targets do not transfer. A hand-picked kinematic descriptor, the stroke's initial direction, matches PMv's held-out geometry better still ($0.65$), so we do not claim structure beyond kinematics; and a label-shuffled PMv control scores $0.41$ (\cref{sec:results}).
  \item \textbf{A validated path to manipulation.} Our evaluation reproduces the published CALVIN performance of Moto's released policy, our one-GPU fine-tuning reproduces its reported low-data result (whose sampling and fine-tuning length Moto does not publish), and we fix the downstream test and its decision rules before running it (\cref{sec:robot}).
\end{itemize}

%% file: sections/03_vocabulary.tex
\section{The Action Vocabulary Is Unstable}
\label{sec:vocab}

\paragraph{The tokenizer.} We study the latent motion tokenizer of Moto~\cite{chen2025moto}, an open-source latent-action model trained on CALVIN~\cite{mees2022calvin}. It reads a pair of frames $(x_t, x_{t+\Delta})$, summarizes their difference with eight query tokens, and quantizes each token to its nearest entry in a codebook of $K{=}128$ vectors of dimension 32~\cite{vandenoord2017neural}. A decoder reconstructs $x_{t+\Delta}$ from $x_t$ and the eight codes, and the reconstruction error is the training loss. In Moto's policy the codes are the targets of video pretraining and the inputs from which actions are decoded, so the codebook acts as the policy's action vocabulary. We train the tokenizer from random initialization for 150k steps in fp32, as in Moto, on a single A100 (\cref{app:train}).

\paragraph{Two seeds, two unrelated vocabularies.} We trained the tokenizer twice with Moto's objective alone, from different random seeds, and encoded the same 1{,}280 CALVIN validation clips (10{,}240 tokens) with both (seed A after 150k steps, seed B after 99k; \cref{app:repro}). The two assignments are nearly unrelated (\cref{tab:vocab}). Their adjusted Rand index, which ignores how codes are labeled, is $0.001$ (a random pairing of the same tokens gives $0.0000 \pm 0.0001$). After the best one-to-one relabeling of codes, they agree on 4.1\% of tokens, against 3.9\% for a random pairing. Both checkpoints reconstruct frames well (mean squared error $0.008$ for seed A and $0.011$ for seed B), so good reconstruction is reached by nearly unrelated assignments. Two caveats apply: the runs differ in training length and seed B changed GPU type midway (\cref{app:repro}), and token-level disagreement does not show that the full eight-token codes carry unrelated motion information; it shows that the objective does not fix the partition.

\paragraph{Precision rewrites the vocabulary.} Limited precision is known to change the discrete outputs of language models~\cite{yuan2025understanding}; here it rewrites the action vocabulary. Evaluating Moto's released tokenizer on identical inputs in bf16 instead of fp32 changes 74\% of its token assignments; TF32 changes 21\%. A randomly initialized tokenizer changes only 2\% under bf16, so the fragility is a property of the trained solution, which places many inputs close to the boundary between codes. Training for 200 steps from the same initialization on the same batches in bf16 and in fp32 ends at nearly the same reconstruction loss ($0.145$ and $0.147$) while assigning only 46\% of tokens alike.

\begin{table}[t]
  \centering
  \small
  \caption{\textbf{The partition of motions into codes is not determined by the objective.} Agreement between two token assignments of the same inputs. Chance agreement for a 128-code vocabulary is about 0.8\%; an adjusted Rand index (ARI) of 0 is the value for unrelated partitions. Seed comparisons use 10{,}240 tokens from 1{,}280 CALVIN validation clips (seed A at 150k steps, seed B at 99k); precision comparisons use 2{,}048 tokens.}
  \label{tab:vocab}
  \begin{tabularx}{\textwidth}{@{}Ycp{5.0cm}@{}}
    \toprule
    Comparison & Same model? & Agreement \\
    \midrule
    Seed A vs.\ seed B, Moto's objective only & no & ARI $0.001$; 4.1\% after best relabeling (random pairing 3.9\%) \\
    Released tokenizer, bf16 vs.\ fp32 evaluation & yes & 26\% same code (74\% reassigned) \\
    Released tokenizer, TF32 vs.\ fp32 evaluation & yes & 79\% same code (21\% reassigned) \\
    Random-init tokenizer, bf16 vs.\ fp32 evaluation & yes & 98\% same code \\
    200 training steps, bf16 vs.\ fp32 (same init, same batches) & --- & 46\% same code; loss $0.145$ vs.\ $0.147$ \\
    \bottomrule
  \end{tabularx}
\end{table}

\paragraph{Why it matters.} The objective fixes how well frames are reconstructed but leaves open which motions share a code. That choice is not cosmetic: the policy decodes actions from the codes, and in an open-loop check, perturbing a trained policy's codes through bf16 moved its predicted arm actions by 60\% of their magnitude, though bf16 also changes the decoder itself (\cref{app:openloop}). We take the freedom as an opportunity. If many vocabularies serve reconstruction equally well, a prior can choose among them, and a measured prior is one we can test.

%% file: sections/04_method.tex
\section{Imposing Premotor Geometry on the Vocabulary}
\label{sec:method}

Our method has three parts: a neural target that says which drawn shapes premotor cortex treats as similar (\cref{sec:target}); a loss that pulls the tokenizer's \emph{code assignment} toward that similarity structure while it keeps learning from robot video (\cref{sec:loss}); and a test that asks whether the imposed structure reaches shapes the loss never saw, relative to matched controls (\cref{sec:eval}).

\subsection{Neural target: premotor geometry of drawing}
\label{sec:target}

We use recordings from ventral premotor cortex of one macaque monkey drawing shapes~\cite{tian2026neural}: four sessions with 49, 40, 55, and 52 PMv units and 307, 241, 348, and 399 strokes. We group strokes by their shape label, keep categories with at least four strokes (19, 18, 24, and 26 per session), and summarize each stroke by the population response from 0.6\,s before to 0.2\,s after stroke onset, projected onto its first eight principal components. The target for a session is its representational dissimilarity matrix (RDM)~\cite{kriegeskorte2008representational}: for every pair of categories, the Euclidean distance between the mean response of one category in a random half of its strokes and that of the other in the complementary half, averaged over 25 splits. Unlike cross-validated distances, this estimator is still inflated by noise, more so for categories with few strokes (\cref{app:neural}).

Because the four sessions record different units, we never pool them: each session contributes its own RDM, and every score is computed per session and then averaged. The target is reliable. The split-half reliability of the RDMs (the noise ceiling) is $0.95$ over all categories and $0.91$--$0.94$ over the five held-out categories, and RDMs of different sessions correlate at $r{=}0.92$ over the 18 categories they share.

\subsection{Aligning the code assignment}
\label{sec:loss}

\paragraph{Two streams, one tokenizer.} The tokenizer trains on two kinds of input that share no examples, following the dual-stream design of \citet{li2019learning}. On CALVIN video it minimizes Moto's reconstruction loss, unchanged. On monkey drawings it encodes renderings of each stroke: the pen trace drawn so far, shown at four points along the stroke and fed as four frame pairs in the same format as robot frames. We describe a stroke by its \emph{code usage}: the fraction of its $4 \times 8$ tokens assigned to each of the 128 codes.

\paragraph{The loss acts on the assignment.} The policy only ever sees which code each input selects, so the alignment must change that choice, not just move continuous features. A loss on a soft relaxation of the assignment can be satisfied by small changes in near-uniform soft assignments without moving any input to a different code. We therefore use a straight-through estimator~\cite{bengio2013estimating}: the forward pass uses the hard nearest-code assignment, and gradients flow through a softmax over negative code distances (temperature $1$), the soft assignment of soft-to-hard vector quantization~\cite{agustsson2017soft}.

\paragraph{Objective.} For each session $s$ we average code usage within each of the 13 training categories and compare the result with the PMv RDM through linear centered kernel alignment (CKA)~\cite{kornblith2019similarity}. The full objective is
\begin{equation}
  \mathcal{L} \;=\; \mathcal{L}_{\text{Moto}}^{\text{robot video}} \;+\; \lambda \Big( \tfrac{1}{4}\textstyle\sum_{s=1}^{4} \big[1 - \mathrm{CKA}(G_s, K_s^{\text{PMv}})\big] \;+\; \gamma\, \mathcal{L}_{\text{collapse}} \Big),
\end{equation}
where $G_s$ is the Gram matrix of category code usage and $K_s^{\text{PMv}}$ the PMv kernel. Because 13 categories could be served by a handful of codes, $\mathcal{L}_{\text{collapse}}$ penalizes low entropy of the overall code-usage histogram. We chose $\lambda{=}0.3$ before training from a probe with a non-neural control target and fixed $\gamma{=}0.1$ and the schedule at the same time: $\lambda$ is zero for the first 1{,}000 steps and ramps up linearly over the next 5{,}000. All other settings follow Moto's tokenizer recipe (\cref{app:train}).

\subsection{Controls and evaluation}
\label{sec:eval}

\paragraph{Control targets.} A tokenizer that becomes PMv-like might owe this to the training procedure rather than to the brain. We therefore train four control tokenizers and one no-prior tokenizer with the PMv tokenizer's recipe, seed, and data order, changing only the target; a second no-prior run differs only in its seed (\cref{tab:arms}). Each control is named for the explanation it tests.

\begin{table}[t]
  \centering
  \small
  \caption{\textbf{Tokenizers and the explanation each one tests.} All are trained for 150k steps with the same recipe; the four control targets use the same loss weight, seed, and data order as PMv.}
  \label{tab:arms}
  \begin{tabularx}{\textwidth}{@{}lp{4.3cm}Y@{}}
    \toprule
    Tokenizer & Target RDM & Tests whether transfer comes from\ldots \\
    \midrule
    \textcolor{pmvc}{\textbf{PMv (ours)}} & PMv, per session & --- \\
    \textcolor{shufc}{\textbf{Shuffled PMv}} & PMv with category labels permuted (over all categories; \cref{app:neural}) & PMv's distance statistics, not which shape has which (imperfect, \cref{sec:results}) \\
    \textcolor{kinc}{\textbf{Kinematics}} & random nonlinear encoders of pen displacement & the motion of the pen \\
    Language & text embeddings of shape descriptions & the shapes' names or semantics \\
    Random & random matrix with PMv's spectrum & imposing any structured target \\
    No prior (2 seeds) & none ($\lambda{=}0$) & training on robot video alone \\
    Released Moto & none (Moto's checkpoint) & the off-the-shelf model \\
    \bottomrule
  \end{tabularx}
\end{table}

\paragraph{Held-out shapes.} Before training, we held out two shape families, circles and L-shaped strokes in four orientations: five categories per session that never enter the loss. Alignment with PMv on these categories is the measure of transfer. For each tokenizer and session we form the RDM of category-mean code usage over the held-out categories, compare it with the PMv RDM on the same categories by Spearman correlation (representational similarity analysis, RSA), and average over the four sessions. Scores use hard code assignments, and they are identical across every GPU type we scored on (\cref{app:repro}).

\paragraph{A kinematic yardstick.} Drawing a shape is a movement, so kinematics alone predicts much of PMv's geometry. We measure how much with a yardstick that needs no training: 100 random nonlinear encoders of the pen's displacement (random Fourier features at five bandwidths, 20 seeds each), scored against PMv exactly like a tokenizer, on the same held-out categories. A tokenizer that beats the yardstick's 95th percentile (0.42, averaged over sessions) captures held-out PMv geometry that 95\% of these random path encoders do not. The yardstick is untrained, summarizes the whole path, and carries no timing, so it is not the strongest kinematic model; we examine a stronger one in \cref{sec:results}. \Cref{app:threshold} gives its per-session values.

\paragraph{Read-out.} Following a rule we fixed before scoring the final checkpoints, the result is the score of the final (150k-step) checkpoint, with the 130k and 140k checkpoints reported alongside it; \cref{fig:results}B and \cref{tab:snapshots} show every checkpoint we scored.

%% file: sections/05_results.tex
\section{Results}
\label{sec:results}

\begin{figure}[t]
  \centering
  \includegraphics[width=\textwidth]{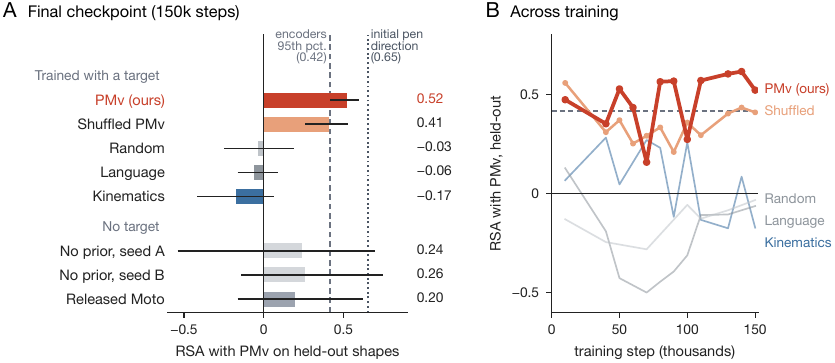}
  \input{captions/fig2}
  \label{fig:results}
\end{figure}

We trained the PMv tokenizer, the four controls, and two no-prior runs for 150k steps each (\cref{tab:arms}) and scored all of them, plus Moto's released tokenizer, on the five held-out categories. \Cref{fig:results} shows the scores and \cref{tab:main} collects them. Every tokenizer is a single training run, and the four sessions come from one animal, so we report every session rather than test statistics.

\subsection{Does premotor geometry reach held-out categories?}

\para{In one training run, the PMv tokenizer reaches 0.52 RSA with PMv on held-out categories, above the released and no-prior tokenizers on average.} The released Moto tokenizer scores $0.20$, and the two no-prior runs score $0.24$ and $0.26$. The PMv tokenizer scores $0.52$, with all four sessions between $0.42$ and $0.60$ (\cref{fig:results}A). The advantage holds on average, not in every session: the no-prior run with the same seed scores higher than the PMv tokenizer in two of four sessions. Random kinematic encoders average $0.26$, and none of the 100 reaches $0.52$ (their best four-session mean is $0.43$). The PMv tokenizer clears their session-averaged 95th percentile ($0.42$; \cref{app:threshold}).

\para{Its scores vary less across sessions.} Their standard deviation is $0.08$, against $0.33$ to $0.55$ for the released tokenizer and the no-prior runs, whose sessions range from $-0.54$ to $0.75$. Single-session scores are noisy: each is a rank correlation over ten distances, and the reliability of PMv's own held-out geometry is $0.91$ to $0.94$.

\begin{table}[t]
  \centering
  \footnotesize
  \caption{\textbf{Held-out alignment and cost at the final checkpoint.} RSA: mean $\pm$ s.d.\ over the four sessions of one training run, on the five held-out categories. ``Sessions $>$ encoders'' counts sessions above that session's 95th percentile of 100 random kinematic encoders (\cref{app:threshold}). Reconstruction error is relative to the mean of the two no-prior runs ($0.0088$ MSE). $^\S$Computed from the training log with the same definition.}
  \label{tab:main}
  \begin{tabular}{@{}lccc@{}}
    \toprule
    Tokenizer or reference & Held-out RSA & Sessions $>$ encoders & Reconstruction error \\
    \midrule
    \textcolor{pmvc}{\textbf{PMv (ours)}} & $\mathbf{0.52 \pm 0.08}$ & \textbf{3/4} & $1.76\times$ \\
    \textcolor{shufc}{\textbf{Shuffled PMv}} & $0.41 \pm 0.12$ & 2/4 & $1.46\times$ \\
    Random & $-0.03 \pm 0.18$ & 0/4 & $1.44\times^\S$ \\
    Language & $-0.06 \pm 0.11$ & 0/4 & $1.35\times$ \\
    \textcolor{kinc}{\textbf{Kinematics}} & $-0.17 \pm 0.22$ & 0/4 & $2.05\times$ \\
    \midrule
    No prior, seed A & $0.24 \pm 0.55$ & 2/4 & $0.88\times$ \\
    No prior, seed B & $0.26 \pm 0.36$ & 1/4 & $1.13\times$ \\
    Released Moto & $0.20 \pm 0.33$ & 1/4 & --- \\
    \midrule
    Initial pen direction (hand-picked) & $0.65 \pm 0.14$ & 4/4 & --- \\
    100 random kinematic encoders & $0.26$ (95th pct.\ $0.42$) & --- & --- \\
    \bottomrule
  \end{tabular}
\end{table}

\subsection{Is it the brain, or the structure of the target?}

\para{Non-neural targets are learned but do not transfer.} The kinematic, language, and random tokenizers were trained with the same loss, weight, seed, and data as the PMv tokenizer, and all three end \emph{below} the no-prior runs: $-0.17$, $-0.06$, and $-0.03$. They did learn their targets: on the training categories every targeted tokenizer matches its own target closely (CKA $0.94$--$0.97$, mean of the last five logged values). What differs is what reaches new categories.

\para{A hand-picked kinematic descriptor matches PMv better than the PMv tokenizer does.} The untrained direction of the first quarter of each stroke reaches $0.65$ held-out RSA with PMv and exceeds the encoders' 95th percentile in all four sessions; using only the first tenth of the stroke, it reaches $0.56$. This is consistent with what the recordings measure. The neural window ends $0.2$\,s after stroke onset while a median stroke lasts 0.8 to 1.0\,s, so PMv's held-out geometry is dominated by how a stroke begins. Our kinematic target averaged random encoders over the whole path, so it is not the strongest kinematic hypothesis, and we did not train a tokenizer toward initial direction. Our results show that training toward PMv gives held-out structure that random path encoders and our kinematic, language, and random targets do not; they do not show structure beyond kinematics.

\para{The shuffled-PMv control is imperfect, and the gap to it is unresolved.} We built the shuffled target by permuting PMv's category labels over all categories of a session before selecting the 13 training rows, so its training block contains the neural patterns of three to five held-out categories per session (\cref{app:neural}). Its score of $0.41$ may therefore reflect the distance statistics of PMv's geometry, relations of held-out categories leaked into training, or seed noise, which one permutation and one seed cannot separate. The PMv tokenizer leads it by $0.11$ at the final checkpoint, about 1.2 times the paired standard error over sessions ($0.09$; per-session differences $-0.11$ to $+0.34$), and in 10 of the 12 session--checkpoint pairs covered by our read-out rule (130k--150k).

\subsection{Is the effect stable across training?}

\para{PMv leads every scored control at each of the last four scored checkpoints (110k--150k).} Single checkpoints are noisy (\cref{fig:results}B), so we fixed in advance that the final checkpoint decides and that the last three are also reported. Over the 130k, 140k, and 150k checkpoints the PMv tokenizer scores $0.60$, $0.62$, and $0.52$ (mean $0.58$). The shuffled tokenizer scores $0.40$, $0.43$, and $0.41$, a mean exactly equal to the threshold ($0.415$). The kinematic tokenizer scores $-0.18$, $0.08$, and $-0.17$. At 150k steps, every control other than shuffled PMv is below zero.

\subsection{What does alignment cost?}

\begin{figure}[t]
  \centering
  \includegraphics[width=\textwidth]{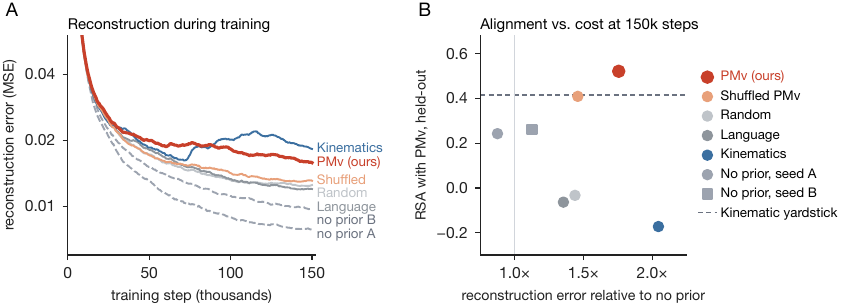}
  \input{captions/fig3}
  \label{fig:cost}
\end{figure}

\para{Every targeted tokenizer pays in reconstruction.} At the final checkpoint the PMv tokenizer reconstructs frames with $1.76\times$ the error of the no-prior runs ($0.0155$ against $0.0088$ MSE; \cref{fig:cost}), the second-highest cost after kinematics ($2.05\times$). The other targets pay $1.35$--$1.46\times$, so much of the cost comes from imposing any target at this weight. The two no-prior runs themselves differ by 29\%, and the PMv tokenizer exceeds both. The alignment also concentrates the vocabulary: on CALVIN validation clips the PMv tokenizer uses 44 of its 128 codes, against 126 for a no-prior run. Whether the cost hurts control is a question for the downstream test (\cref{sec:robot}).

%% file: captions/fig2.tex
\caption{\textbf{Premotor geometry transfers to held-out shapes.} \textbf{(A)}~RSA between each tokenizer's code geometry and PMv on the five held-out shape categories, at the final checkpoint. Bars: mean over the four recording sessions of one training run; black lines: range over sessions; numbers: means (s.d.\ in \cref{tab:main}). Vertical lines are untrained references. Dashed ($0.42$): the kinematic yardstick, each session's 95th percentile of 100 random nonlinear encoders of pen motion (mean $0.26$), averaged over sessions (\cref{app:threshold}); only the PMv tokenizer exceeds it. The 95th percentile of the encoders' four-session means is lower ($0.39$), and the shuffled tokenizer also exceeds that. Dotted ($0.65$): a hand-picked kinematic descriptor, the stroke's initial direction, which exceeds every tokenizer. \textbf{(B)}~The same score at every scored checkpoint (mean over sessions); PMv and shuffled PMv are emphasized, and the dashed line is the same yardstick. Single checkpoints are noisy because each session's score is a rank correlation over ten distances, which is why the final checkpoint was fixed as the read-out in advance.}

%% file: captions/fig3.tex
\caption{\textbf{Alignment costs reconstruction.} \textbf{(A)}~Training reconstruction error (running mean over 5k steps); dashed lines are the two no-prior runs, and labels give each line's tokenizer. \textbf{(B)}~Held-out RSA with PMv (dashed: kinematic yardstick) against final reconstruction error (mean of the last five logged values, relative to the no-prior mean of $0.0088$). Every targeted tokenizer raises reconstruction error, to $1.35$--$2.05\times$ the no-prior level; the PMv tokenizer reaches the highest alignment, at the second-highest cost.}

%% file: sections/06_robot.tex
\section{Toward Manipulation: A Validated, Pre-Specified Downstream Test}
\label{sec:robot}

Alignment with PMv is a means, not an end; the question that matters for robotics is whether the aligned vocabulary changes what a policy can do. Answering it requires training Moto's full policy on each tokenizer, which we have built, validated, and launched. This section reports the validation and fixes the test before any of its numbers exist.

\paragraph{Pipeline.} Moto trains a policy in two further stages on top of the tokenizer~\cite{chen2025moto}. \emph{Stage~2} pretrains a GPT-style model to predict the tokenizer's latent-action tokens from CALVIN video and language instructions; the tokenizer is frozen, so its vocabulary is the prediction target. \emph{Stage~3} fine-tunes the model to output robot actions. We follow Moto's low-data setting on CALVIN ABC$\rightarrow$D~\cite{mees2022calvin}: Stage~3 sees 1\% of the labeled training episodes (179 of 17{,}870, drawn once with a fixed seed and shared by every tokenizer), while Stages~1--2 use all unlabeled video. Policies are evaluated on CALVIN's 1{,}000 fixed instruction chains of five tasks each.

\paragraph{Validation.} Before training on any of our tokenizers, we checked that the pipeline reproduces published numbers (\cref{tab:calvin}). Evaluating Moto's released policy, our setup completes $3.13 \pm 0.06$ tasks in a row on average against the published $3.10$, with per-step success rates within two points of the paper. For the low-data setting, we fine-tuned Moto's released Stage-2 model on our 1\% split with our training code: it solves the first task of a chain 50.7\% of the time, against the 52.5\% success rate Moto reports with 1\% of labeled data; Moto does not publish how its 1\% was drawn or how long it fine-tuned, so this comparison is approximate. These checks validate evaluation and low-data fine-tuning; our Stage-2 recipe is validated by its matched schedule and resumption test (\cref{app:recipe}), not yet by an end-to-end result.

\begin{table}[t]
  \centering
  \footnotesize
  \caption{\textbf{Our CALVIN pipeline reproduces Moto.} Success rate at each step of 1{,}000 five-task chains (ABC$\rightarrow$D) and the average number of consecutive tasks completed (Avg.\ Len., $\pm$ s.e.). $^\dagger$Moto reports a 52.5\% ``success rate'' with 1\% of labeled data without specifying the step~\cite{chen2025moto}; we list it under the first task.}
  \label{tab:calvin}
  \begin{tabular}{@{}llcccccc@{}}
    \toprule
    Labeled data & Policy & 1 & 2 & 3 & 4 & 5 & Avg.\ Len. \\
    \midrule
    \multirow{2}{*}{100\%} & Published~\cite{chen2025moto} & 0.897 & 0.729 & 0.601 & 0.484 & 0.386 & 3.10 \\
                           & Released checkpoint, our evaluation & 0.899 & 0.748 & 0.610 & 0.491 & 0.380 & $3.13 \pm 0.06$ \\
    \midrule
    \multirow{2}{*}{1\%}   & Published~\cite{chen2025moto} & 0.525$^\dagger$ & -- & -- & -- & -- & -- \\
                           & Released Stage~2 + our 1\% Stage~3 & 0.507 & 0.206 & 0.071 & 0.027 & 0.005 & $0.82 \pm 0.03$ \\
    \bottomrule
  \end{tabular}
\end{table}

\paragraph{A faithful one-GPU recipe.} Moto's released training loop applies only the last micro-batch of each gradient-accumulation window, so its updates used 128 samples, not the nominal 512. We reproduce this effective recipe on one GPU and add exact-state resumption (\cref{app:recipe}); skipping the samples Moto discarded cuts Stage~2 to a projected 1.8 days per tokenizer, from the measured step time.

\paragraph{The test, fixed in advance.} Four tokenizers enter Stages~2--3 with identical seeds, data order, and hardware: the two runs without a prior, the PMv tokenizer, and the kinematic control. We compare policies chain by chain on the same 1{,}000 chains, with 95\% bootstrap confidence intervals over chains and a Wilcoxon signed-rank test. PMv \emph{improves manipulation} only if its gain over \emph{each} no-prior run has a confidence interval above zero and exceeds the gap between the two no-prior runs. If PMv passes, it must also beat the kinematic tokenizer to show an advantage specific to neural data; if it does not, a neural-specific advantage is not demonstrated. Confidence intervals over chains capture evaluation noise, not variation across training seeds, so one seed per tokenizer bounds what this test can show. The four Stage-2 runs are in progress, and we will add their results to this paper in a revision.

%% file: sections/02_related.tex
\section{Related Work}
\label{sec:related}

\para{Latent actions from video.} Inferring latent actions from unlabeled observation dates to imitation from observation~\cite{edwards2019imitating}. It has since been scaled to generative interactive environments~\cite{bruce2024genie}, latent action pretraining~\cite{schmidt2024learning,ye2025latent}, and vision-language-action policies built on discrete latent motion tokens~\cite{chen2025moto,bu2025learning}. A growing literature examines what these latent actions capture: they can absorb distractors unless supervised~\cite{nikulin2025latent}, reconstruction training can leave them poorly aligned with language~\cite{li2026lost}, and compositional structure can be built into them~\cite{wei2026learning}. Closest to our first finding, \citet{wen2026algebraic} show that algebraic-consistency metrics do not certify temporal structure in latent-action models and call for seed-budget analyses, and \citet{schur2026identifying} prove that latent actions are identifiable up to relabeling when demonstrators are diverse and tagged. Moto's data carries no demonstrator tags, and the adjusted Rand index we report is invariant to relabeling, so our seed result is consistent with theirs. We add a direct measurement for a released model and use the freedom it reveals to impose a measured neural geometry, rather than adding action supervision.

\para{Underspecification.} Networks trained identically up to the random seed can learn different internal representations~\cite{li2016convergent,mehrer2020individual}. More generally, many solutions perform equally well on the training objective in modern machine learning~\cite{damour2022underspecification}, and disentangled representations are not identifiable without inductive biases~\cite{locatello2019challenging}. Limited-precision arithmetic is known to change the discrete outputs of language models~\cite{yuan2025understanding}. We find that both seed and precision change the action vocabulary of a latent-action model.

\para{Brain-guided machine learning.} Similarity between models and brains has been measured extensively in vision~\cite{yamins2014performance,yamins2016using,schrimpf2018brainscore,schrimpf2020integrative}; \citet{sucholutsky2025getting} review representational alignment. A smaller body of work imposes neural structure during training, through similarity losses toward neural recordings~\cite{li2019learning,fong2018using,federer2020improved,safarani2021towards,pirlot2022improving}, a fixed model of primary visual cortex~\cite{dapello2020simulating}, or alignment to macaque inferior temporal cortex~\cite{dapello2023aligning}. \citet{federer2020improved} found that random targets with the statistics of neural data can also help, and \citet{attias2025pixel} that pixel-based similarities can stand in for neural data. These are the reasons we include random and shuffled-neural targets as controls.

On the action side, goal-driven models relate trained networks to motor-system recordings~\cite{sussillo2015neural,michaels2020goal,aldarondo2024virtual,almani2024musim}. Relational Geometry Forcing transfers the pairwise feature geometry of a frozen 3D foundation model into world-action models~\cite{xu2026relational}, a non-neural relational prior close to ours in the form of its loss. \citet{oota2026brain} relate vision-language and large-action (GUI-agent) models to human brain activity during gameplay.

\para{Concurrent work.} Closest to ours, \citet{kapitonova2026distilling} align the hidden states of recurrent and transformer controllers with motor-cortex population geometry recorded during reaching. They compare against kinematic and shuffled teachers, and they improve reaching by a simulated arm. We differ in four ways: our target is premotor cortex during drawing rather than motor cortex during reaching; we constrain the discrete code assignment of a latent-action model trained on video rather than the continuous states of a controller; our neural data and robot data share no conditions, hence the two streams; and our downstream benchmark is manipulation on CALVIN.

\para{Premotor cortex and representational similarity.} Ventral premotor cortex encodes goal-directed hand actions~\cite{rizzolatti1988functional}, and during drawing it represents strokes as reusable, categorical action symbols~\cite{tian2026neural}, which is why we chose it as the target for an action vocabulary. Our analyses follow representational similarity analysis~\cite{kriegeskorte2008representational,nili2014toolbox} with distances estimated across independent halves of the data~\cite{walther2016reliability}, and our training loss uses centered kernel alignment~\cite{kornblith2019similarity}.

%% file: sections/07_limitations.tex
\section{Limitations}
\label{sec:limits}

\paragraph{Not beyond kinematics.} The stroke's initial direction, an untrained descriptor, matches PMv's held-out geometry better ($0.65$) than the PMv tokenizer ($0.52$), and no tokenizer was trained toward it, so the strongest kinematic hypothesis is untested (\cref{sec:results}).

\paragraph{One run per target, one animal.} Each tokenizer was trained once, and the four sessions share one monkey and most of their categories, so they are not independent replicates. The PMv margin over the shuffled control ($0.11$, about 1.2 paired standard errors) is therefore unresolved, and that control is itself imperfect (\cref{sec:results}).

\paragraph{A small held-out set.} Each session's held-out RDM has five categories, four orientations of an L-shaped stroke and a circle, so each score is a rank correlation over ten distances. PMv does not set the circle apart (a model that only separates the circle from the L shapes scores $0.07$), so the test mostly probes how PMv arranges four orientations of one shape. On this set, image identity and drawn category are confounded.

\paragraph{Neural preprocessing saw all categories.} The principal components that summarize each PMv response were fit on all strokes, including those of held-out categories. The loss never saw the held-out categories, but the neural space they are measured in was not fit on training data alone.

\paragraph{Held-out categories, not held-out sessions.} The loss used all four recording sessions, so our results concern new shapes, not a new recording.

\paragraph{All alignment evidence comes from strokes.} Alignment is measured on monkey drawings rendered as frame pairs. Whether the PMv-aligned code geometry also organizes the robot's own motions is what the downstream test probes.

\paragraph{No manipulation results yet.} This version shows, in one run per target, that premotor geometry can be imposed on a robot's action vocabulary and reaches held-out shapes. It does not yet show that doing so helps a robot; \cref{sec:robot} specifies how we will test that.

%% file: sections/08_conclusion.tex
\section{Conclusion}
\label{sec:conclusion}

In Moto's latent-action tokenizer, the objective does not pin down which motions share a code: a different seed or a lower numerical precision rewrites most of the assignment while reconstruction stays good. We used that freedom to impose a measured neural geometry. Trained toward macaque premotor cortex, the tokenizer aligns with PMv on five held-out shape categories in one training run. It does so beyond what kinematic, language, or random targets produce and beyond any of 100 random kinematic encoders, though not beyond a hand-picked kinematic descriptor, the stroke's initial direction.

Several questions remain open: how much of the effect is specific to PMv, whether it survives more seeds and cleaner controls, and whether it helps a robot act. The downstream manipulation test, already running, will answer the last. Vision models have long borrowed structure from the visual cortex; this work takes a measured step toward doing the same for action.

%% file: sections/A_appendix.tex
\section*{Appendix}
\addcontentsline{toc}{section}{Appendix}

\section{The Kinematic Yardstick}
\label{app:threshold}

The yardstick is the 95th percentile of held-out RSA over 100 random nonlinear encoders of pen displacement (random Fourier features at five bandwidths, 20 seeds each), computed per recording session on the five held-out categories and averaged over sessions (\cref{tab:threshold}). We computed it after training the tokenizers. The PMv tokenizer ($0.52$) clears it; the shuffled ($0.409$), random, language, and kinematic tokenizers do not. Two details matter for the shuffled control. The mean of its last three checkpoints is $0.415$, equal to the yardstick. And because the yardstick averages four per-session 95th percentiles, it differs from the 95th percentile of the encoders' four-session \emph{means}, which is lower ($0.391$); the shuffled tokenizer exceeds that too.

\begin{table}[h]
  \centering
  \small
  \caption{\textbf{The kinematic yardstick.} 95th percentile of 100 random kinematic encoders on the five held-out categories, per recording session, and their average.}
  \label{tab:threshold}
  \begin{tabular}{@{}ccccc@{}}
    \toprule
    Session 1 & Session 2 & Session 3 & Session 4 & Yardstick \\
    \midrule
    0.479 & 0.576 & 0.358 & 0.248 & 0.415 \\
    \bottomrule
  \end{tabular}
\end{table}

\section{Neural Data and Targets}
\label{app:neural}

\para{Recordings.} Ventral premotor cortex, one monkey, four sessions, from \citet{tian2026neural} (CC BY 4.0). Units: 49, 40, 55, 52. Strokes: 307, 241, 348, 399. Shape categories with at least four strokes: 19, 18, 24, 26. Each stroke's response is the mean firing rate from $-0.6$\,s to $+0.2$\,s around stroke onset, projected onto the first eight principal components; the principal components are fit on all strokes, including held-out ones.

\para{RDMs.} For each pair of categories, the distance is the Euclidean distance between the mean response of one category in a random half of its strokes and the mean of the other category in the complementary half, averaged over 25 random splits. Halves are split by stroke, not by trial. Unlike cross-validated (crossnobis) distances, this estimator remains inflated by noise, more so for categories with few strokes. Its RDMs nevertheless agree closely with crossnobis RDMs (Spearman $0.9975$; kernel correlation $0.951$). Split-half reliability (Spearman--Brown):
\begin{itemize}
  \item all categories: $0.955$, $0.960$, $0.960$, $0.944$;
  \item five held-out categories: $0.914$, $0.942$, $0.935$, $0.907$.
\end{itemize}
RDMs of different sessions correlate at $r{=}0.925$ over the 18 categories they share.

\para{Confounds in the stroke set.} About half of the held-out strokes follow another stroke within the same trial. Grid locations are not balanced across categories, and only the tokenizer, not the PMv RDM, sees absolute position on the canvas.

\para{Split.} Held-out families: circles and L-shaped strokes, the latter in four orientations (five categories). Training categories (13): two V-shapes, three arcs, three squiggles, two U-squares, and three square zigzags. The split was frozen before training.

\para{Control targets.}
\begin{itemize}
  \item \textbf{Kinematics:} the average over the random kinematic encoders of their category RDM, per session.
  \item \textbf{Shuffled PMv:} the PMv kernel with rows and columns permuted by a derangement \emph{over all categories of the session}; the 13 training rows were selected afterwards. The training block therefore contains the neural patterns of 4, 5, 4, and 3 held-out categories in the four sessions (all five in session 2). This leaks held-out relations into the shuffled control's training target, so the control does not cleanly isolate PMv's distance statistics.
  \item \textbf{Random:} a random positive semi-definite matrix with the PMv kernel's eigenvalue spectrum.
  \item \textbf{Language:} Euclidean distances between L2-normalized t5-base embeddings (Moto's own text encoder) of short descriptions generated from the shape labels. In each session two to four category pairs share an identical description and therefore have distance zero.
\end{itemize}
Only the shuffled target directly includes held-out category relations. Separately, the neural PCA behind PMv-derived targets and evaluation RDMs was fit on all categories (\cref{sec:limits}).

\section{Training Details}
\label{app:train}

\para{Tokenizer.} Moto's latent motion tokenizer with its released architecture (121.2M trainable parameters; frozen MAE-Large image encoder~\cite{he2022masked}). Moto trains it for 150k steps with batch 256 on four GPUs. We train for 150k steps with micro-batch 32 and gradient accumulation 8 (batch 256) on one A100 in fp32, from random initialization, with Moto's optimizer and learning-rate schedule. Batch contents are a deterministic function of the seed and step, so tokenizers with the same seed see identical robot data. The five targeted tokenizers share seed 0 with the first no-prior run and are identical to it until the alignment loss starts.

\para{Stroke stream.} Each stroke is resampled to 200 points along its arc length and rendered as the growing pen trace at $224 \times 224$ pixels. Four frame pairs spaced along the stroke enter the tokenizer like robot frame pairs. All sessions share one drawing extent, and strokes are not augmented. Each step draws four strokes per training category. The temperature is $1$. $\lambda{=}0.3$ was chosen from a probe that used the random (non-neural) target; $\gamma{=}0.1$ and the schedule were fixed at the same time, before training.

\para{Compute.} On one A100, a tokenizer takes about 330 GPU-hours without a prior and about 520 with one (roughly 8 and 12.5\,s per step of 256). The seven tokenizers in this paper used about 3{,}300 A100-hours.

\section{Reproducibility and Measurement}
\label{app:repro}

\para{Device invariance.} Held-out RSA was identical, to every digit, across four GPU types (A100 80GB PCIe, A100 40GB, A30, A10) for every checkpoint scored on more than one of them. CPU scores are not always identical: for the language tokenizer and the second no-prior run, one session differed on CPU, and we report the GPU values. The random tokenizer's 150k score was computed on CPU and then on an A100, with identical results. Code assignments of \emph{robot} frames are not device-invariant: 2.5--8\% of 10{,}240 assignments differ between A30 and A100, so robot-side comparisons use a single GPU type.

\para{The seed comparison.} \Cref{sec:vocab} compares seed A's 150k checkpoint with seed B's 99k checkpoint, the latest pair whose robot-frame assignments were computed on the same run; a matched 150k comparison is queued.

\para{Hardware and process events.} The second no-prior run moved from an A100 to an A10 and back during training; resuming across these two chip architectures is not bit-exact, so this run contains two numerical discontinuities. The shuffled tokenizer trained for 1{,}000 of its steps on an A100 40GB (same architecture). For about two days, the random tokenizer and a paused second-seed PMv run each ran as two processes continuing from one checkpoint; every saved state is a valid continuation, but these runs cannot be reproduced bit-for-bit from their seeds. Resumption is otherwise bit-exact.

\para{Read-out rule.} The secondary read-out, the mean of the 130k, 140k, and 150k checkpoints, is:
\begin{itemize}
  \item $0.58$ for the PMv tokenizer;
  \item exactly $0.415$ for the shuffled tokenizer;
  \item $-0.09$ for the kinematic tokenizer;
  \item not available for the random and language tokenizers, whose 140k checkpoints were not scored.
\end{itemize}
\Cref{tab:snapshots} lists every checkpoint we scored.

\begin{table}[h]
  \centering
  \scriptsize
  \caption{\textbf{Held-out RSA with PMv at every scored checkpoint} (mean over four sessions; ``--'' = not scored).}
  \label{tab:snapshots}
  \setlength{\tabcolsep}{3.2pt}
  \input{tables/snapshots}
\end{table}

\section{Open-Loop Check: Codes Steer the Policy's Actions}
\label{app:openloop}

To check that a policy's actions depend on its codes, we evaluated Moto's released policy on 256 CALVIN frame pairs for one instruction (``push the block to the left''), once in fp32 and once in bf16. bf16 changed 74\% of the latent-action codes the policy predicted, and its predicted arm actions moved by 60\% of their mean magnitude; replacing the codes with random ones moved them by 133\%. This is a single open-loop check on one checkpoint and one instruction, not a measure of task success. It also does not isolate the codes: bf16 changes the action decoder's arithmetic as well, so part of the change in actions may come from pathways other than the codes. A causal test would swap codes while holding the decoder and its precision fixed.

\section{A Faithful One-GPU Recipe for Moto's Policy}
\label{app:recipe}

\para{The accumulation behavior.} Moto's policy trainer calls \texttt{optimizer.zero\_grad()} at the start of each micro-batch inside \texttt{accelerate}'s gradient-accumulation context. With the pinned \texttt{accelerate} version (1.0.1), \texttt{zero\_grad} takes effect only on the micro-batch that completes an accumulation window. It therefore clears the gradient accumulated over the window just before that micro-batch's backward pass. We confirmed this on a toy model: the applied update equals the last micro-batch's gradient exactly. Moto's released Stage-2 run (eight GPUs, micro-batch 16, accumulation 4) therefore updated on 128 samples per step while drawing 512. We have not confirmed this with Moto's authors.

\para{Our recipe.} We reproduce the effective recipe on one GPU:
\begin{itemize}
  \item each update accumulates correctly over eight micro-batches of 16 (128 samples);
  \item each epoch uses the quarter of the data whose gradients Moto's run used;
  \item the steps per epoch and the cosine learning-rate schedule match Moto's in its own units.
\end{itemize}
Not computing the discarded samples makes Stage~2 about four times cheaper. Our runs train at about 7.5\,s per step, which projects to 1.8 days per tokenizer on one A100. We use the same 128-sample update for Stage~3; Moto's Stage~3 ran on four GPUs, which corresponds to 64 samples per update. Our low-data check (\cref{tab:calvin}) used our setting.

\para{Resumption and data order.} Our trainer saves and restores the optimizer, learning-rate schedule, random-number state, counters, and data position. It draws each epoch's data order from a seeded permutation that is identical for every tokenizer. A run stopped and resumed after 200 of 400 micro-batches matches an uninterrupted run to $8 \times 10^{-4}$ relative loss, with an identical learning rate.

\section{Downstream Protocol, Fixed Before Any Result}
\label{app:protocol}

\para{Setup.}
\begin{itemize}
  \item \textbf{Tokenizers:} both no-prior runs, the PMv tokenizer, and the kinematic tokenizer, each at 150k steps, exported unchanged.
  \item \textbf{Stage~2:} Moto's CALVIN configuration (10 of 20 cosine epochs), fp32, seed 0 for all four.
  \item \textbf{Stage~3:} the fixed 1\% split (179 episodes), initialized from each tokenizer's own Stage-2 checkpoint, 20 epochs.
  \item \textbf{Evaluation:} Moto's evaluation code, unchanged, on CALVIN's 1{,}000 ABC$\rightarrow$D chains with seed 12345, on A100 GPUs only. We split the chains across eight processes, as in Moto's multi-GPU evaluation; this changes the sampling random stream relative to a single-process run.
  \item \textbf{Metric:} the average number of consecutive tasks completed, with per-step success also reported.
  \item \textbf{Statistics:} paired over chains, with 95\% bootstrap confidence intervals from 10{,}000 chain resamples and a Wilcoxon signed-rank test.
\end{itemize}

\para{Decision rules.} \textbf{PMv improves manipulation} if, for each no-prior run, the confidence interval of its gain lies above zero and its mean gain exceeds the difference between the two no-prior runs. \textbf{No neural-specific advantage:} if PMv passes but its gain over the kinematic tokenizer has a confidence interval that includes or lies below zero, an advantage specific to premotor cortex is not demonstrated.

%% file: tables/snapshots.tex
\begin{tabular}{@{}lcccccccccccc@{}}
\toprule
Tokenizer & 10k & 40k & 50k & 60k & 70k & 80k & 90k & 100k & 110k & 130k & 140k & 150k \\
\midrule
PMv & $0.47$ & $0.35$ & $0.53$ & $0.43$ & $0.16$ & $0.56$ & $0.57$ & $0.27$ & $0.57$ & $0.60$ & $0.62$ & $0.52$ \\
Shuffled PMv & $0.56$ & $0.31$ & $0.37$ & $0.25$ & $0.29$ & $0.33$ & $0.21$ & $0.36$ & $0.29$ & $0.40$ & $0.43$ & $0.41$ \\
Random & $-0.13$ & $-0.25$ & -- & -- & $-0.28$ & -- & -- & $-0.06$ & $-0.13$ & $-0.08$ & -- & $-0.03$ \\
Language & $0.13$ & $-0.19$ & $-0.43$ & -- & $-0.50$ & $-0.45$ & $-0.39$ & $-0.31$ & $-0.11$ & $-0.11$ & -- & $-0.06$ \\
Kinematics & $0.07$ & $0.28$ & $0.05$ & -- & $0.27$ & $0.23$ & $-0.12$ & $0.26$ & $-0.13$ & $-0.18$ & $0.08$ & $-0.17$ \\
\bottomrule
\end{tabular}